\documentclass[sigconf]{acmart}
\usepackage{multirow}
\usepackage{svg}
\AtBeginDocument{%
  }

\copyrightyear{2025}
\acmYear{2025}
\setcopyright{cc}
\setcctype{by}
\acmConference[BCB '25]{Proceedings of the 16th ACM International Conference on Bioinformatics, Computational Biology, and Health Informatics}{October 11--15, 2025}{Philadelphia, PA, USA}
\acmBooktitle{Proceedings of the 16th ACM International Conference on Bioinformatics, Computational Biology, and Health Informatics (BCB '25), October 11--15, 2025, Philadelphia, PA, USA}
\acmDOI{10.1145/3765612.3767245}
\acmISBN{979-8-4007-2200-4/2025/10}

\begin{document}

\title[HyMaTE]{HyMaTE: A Hybrid Mamba and Transformer Model for EHR Representation Learning}


\author{Md Mozaharul Mottalib}
  \affiliation{
  \institution{University of Delaware}
  \city{Newark}
  \state{Delaware}
  \country{USA}
}
\email{mmmdip@udel.edu}
\author{Thao-Ly T. Phan}
\affiliation{
    \institution{Thomas Jefferson University \& Nemours Children's Health}
    \city{Wilmington}
    \state{Delaware}
    \country{USA}
}
\email{tphan@nemours.org}

\author{Rahmatollah Beheshti}
\affiliation{
  \institution{University of Delaware}
  \city{Newark}
  \state{Delaware}
  \country{USA}
}
\email{rbi@udel.edu}

\renewcommand{\shortauthors}{Mottalib et al.}

\begin{abstract}
Electronic health Records (EHRs) have become a cornerstone in modern-day healthcare. They are a crucial part for analyzing the progression of patient health; however, their complexity, characterized by long, multivariate sequences, sparsity, and missing values-poses significant challenges in traditional deep learning modeling. While Transformer-based models have demonstrated success in modeling EHR data and predicting clinical outcomes, their quadratic computational complexity and limited context length hinder their efficiency and practical applications. On the other hand, State Space Models (SSMs) like Mamba present a promising alternative offering linear-time sequence modeling and improved efficiency for handling long sequences, but focus mostly on mixing sequence-level information rather than channel-level data. To overcome these challenges, we propose HyMaTE (A \underline{Hy}brid \underline{Ma}mba and \underline{T}ransformer Model for \underline{E}HR Representation Learning), a novel hybrid model tailored for representing longitudinal data, combining the strengths of SSMs with advanced attention mechanisms. By testing the model on predictive tasks on multiple clinical datasets, we demonstrate HyMaTE's ability to capture an effective, richer, and more nuanced unified representation of EHR data. Additionally, the interpretability of the outcomes achieved by self-attention illustrates the effectiveness of our model as a scalable and generalizable solution for real-world healthcare applications. Codes are available at: \url{https://github.com/healthylaife/HyMaTE}.
\end{abstract}




\maketitle

\section{Introduction}
Electronic Health Records (EHRs) are a crucial resource for analyzing patient health progression. They document the entire medical histories of patients, including diagnoses, procedures, observations, medications, laboratory tests, demographics, and clinical notes, forming detailed chronological sequences. With over $80\%$ of hospitals in the US and Canada adopting EHR systems, this extensive data offers an unparalleled resource for training advanced models \citep{fallahpour2024ehrmamba}. These models hold the potential to personalize treatment plans, uncover disease patterns, detect the onset of illnesses, and enhance clinical practice.

However, the application of deep learning models to EHR data faces significant challenges due to its inherent characteristics. EHR data is notably complex, characterized by long, multivariate sequences, high sparsity, and numerous missing values \citep{tonekaboniunsupervised}. This data is often high-dimensional and irregularly distributed across time, making it difficult to apply standard time series analysis methods designed for densely sampled data \citep{tipirneni2022self}. The semantic information in time series data, including EHRs, is primarily derived from temporal variation, which adds complexity to its understanding.

While Transformer-based architectures \citep{li2020behrt, pang2021cehr, rasmy2021med, li2022hi, labach2023duett, rupp2023exbehrt} have shown remarkable success in modeling EHR data and predicting clinical outcomes, they suffer from critical limitations. A primary concern is their quadratic computational complexity and insufficient context lengths \citep{fallahpour2024ehrmamba}. These issues can restrict their efficiency and practical deployment, especially when attempting to capture a patient's entire medical history, which can span thousands of tokens. The quadratic scaling of self-attention layers can limit the input sequence length, often necessitating truncation, the use of shallow models, or quantization methods, which can negatively impact accuracy \citep{labach2023duett}. Transformers can also struggle to maintain local sequential coherence, as their design does not inherently prioritize temporal dependencies. Furthermore, applying transformers directly across the time dimension of EHR data can lead to a loss of information along the event dimension, limiting the model's ability to capture relationships between different event types \citep{zhu2025hybrid}.

Recently, State Space Models (SSMs), particularly the Mamba architecture \citep{gumamba}, have emerged as a promising alternative for sequence modeling. Mamba offers linear-time sequence modeling and enhanced efficiency for long sequences, directly addressing issues like low inductive bias and the quadratic complexity inherent in transformers \citep{patro2025simba, fallahpour2024ehrmamba}. This linear scaling of computational and memory requirements makes Mamba a more feasible solution for processing extensive medical histories in resource-limited hospital settings. However, traditional Mamba-based models have primarily focused on sequence mixing, but not channel mixing, which is crucial for multivariate time series data like EHRs \citep{patro2025simba}. 

To address these multifaceted challenges, we propose HyMaTE (A \underline{Hy}brid \underline{Ma}mba and \underline{T}ransformer Model for \underline{E}HR Representation Learning. HyMaTE is a novel hybrid model specifically designed for longitudinal data representation. It strategically combines the strengths of State Space Models (SSMs), particularly Mamba, with advanced attention mechanisms, i.e., the backbone of transformers. This approach is engineered to overcome the computational limitations of transformers and to address the channel-mixing challenge in Mamba, aiming to achieve state-of-the-art performance across various clinical prediction tasks while ensuring efficiency and scalability for real-world healthcare applications.

\section{Related Works}
The analysis of EHRs is a critical area for understanding patient condition progression and disease trajectories, with EHRs serving as a comprehensive record of medical histories, including diagnoses, procedures, observations, medications, and laboratory test results. However, the intrinsic characteristics of EHR data, such as long, multivariate sequences, sparsity, and missing values, present substantial hurdles for traditional deep learning models. Effective modeling requires capturing both temporal relationships and inter-variate dependencies within this complex data.

Initially, Recurrent Neural Networks (RNNs) and their variants, such as Long Short-Term Memory (LSTM) and Gated Recurrent Units (GRU) \citep{chung2014empirical}, demonstrated success in modeling sequential data, including EHR. To accommodate the sparsity and irregularity prevalent in EHR data, researchers proposed various modifications. These included integrating learnable imputations, interpolation of components, binning, and architectural modifications like the GRU with trainable Decays (GRU-D) \citep{ienco2020deep} and Continuous Time-GRU (CT-GRU) \citep{mozer2017discrete}. Despite these advancements, RNN-based models inherently suffer from limitations due to their sequential nature, which restricts parallelization and can lead to issues such as vanishing or exploding gradients, impeding their capacity to learn long-range dependencies effectively.

The advent of Transformer models marked a significant breakthrough in sequence modeling, particularly in Natural Language Processing (NLP), and their success led to widespread adaptation for EHR data analysis and clinical outcome prediction. Transformers utilize self-attention mechanisms to capture long-range temporal dependencies and complex multivariate correlations. Notable Transformer-based models developed for EHR include TransformEHR \citep{rajkomar2018scalable}, BEHRT \citep{li2020behrt}, CEHR-BERT \citep{pang2021cehr}, Med-BERT \citep{rasmy2021med}, RAPT \citep{ren2021rapt}, STraTS \citep{tipirneni2022self}, and DuETT \citep{labach2023duett}, which aim to leverage the rich information in patient records. However, a fundamental challenge with Transformer architectures in the context of EHR is their quadratic computational complexity and memory requirements with respect to input sequence length $(\textbf{O}(N^2))$. This quadratic scaling significantly limits the context window size that can be practically processed, often restricting models to short contexts despite real patient histories spanning tens of thousands of tokens. Additionally, while transformers excel at sequence mixing, their direct application to multivariate time series data like EHRs often overlooks the crucial aspect of channel mixing, which is vital for understanding interactions between different types of variates. Attempts to address these limitations have included hierarchical Transformer architectures (e.g., Hi-BEHRT), modified attention mechanisms (e.g., time-aware attention in RAPT), and input engineering techniques like time binning and triplet embeddings (e.g., DuETT, STraTS). Graph-based networks have been also proposed for sequence modeling. A novel graph attention network framework was proposed that utilizes a hierarchical approach to generate embeddings from EHR \citep{piya2024healthgat}. Also, a model was proposed that leverages temporal visit embeddings extracted from a graph transformer and uses a BERT-based model to obtain more robust patient representations, especially on longer EHR sequences \citep{poulain2024graph}.

State Space Models (SSMs) have emerged as a promising alternative for sequence modeling, particularly offering linear-time efficiency for processing long sequences. The Mamba architecture \citep{gumamba} is a recent advancement in SSMs that directly addresses the limitations of low inductive bias and quadratic complexity often associated with transformers, providing robust sequence modeling with enhanced efficiency. Mamba's unique approach integrates the current input token within its state space, facilitating in-context learning. However, Mamba-based models have traditionally concentrated on sequence mixing rather than channel mixing, which is essential for multivariate time series data like EHRs \citep{patro2025simba}.

To overcome these combined challenges, several hybrid or Mamba-based solutions have been proposed. SiMBA-TS (Simplified Mamba-Based Architecture for Time Series Forecasting) \citep{patro2025simba} addresses Mamba's channel mixing limitation by combining the selective scan SSM (Mamba) for token mixing with a novel Einstein Matrix Multiplication (EinFFT) method for channel mixing, achieving convincing performance in long-term time series forecasting across multiple datasets. EhrMamba \citep{fallahpour2024ehrmamba} is introduced as a robust foundation model for EHR, built on the Mamba architecture. It can process sequences up to $300\%$ longer than previous Transformer-based models due to its linear computational cost. EhrMamba also introduces Multi-task Prompted Finetuning (MPF) to learn multiple clinical tasks simultaneously, and is designed for both EHR forecasting and clinical predictive modeling. Hybrid Transformer-Mamba models such as TransMamba \citep{li2025transmamba, chen2025transmamba} and Nemotron-H \citep{blakeman2025nemotron} explore the synergistic integration of both architectures. These models leverage the Transformer's strengths (e.g., rich feature representations, parallelization) with Mamba's efficiency and powerful sequential modeling capabilities, aiming to achieve enhanced performance and address instability issues. Some innovative designs even allow for dynamic switching between attention and SSM mechanisms at different token lengths and layers to optimize efficiency and performance.

\paragraph{Interpretability in EHR Models}
The inherent complexity of deep learning models often makes it challenging to explain their predictions to clinicians and healthcare providers, despite their high predictive accuracy. This interpretability gap is a significant concern for the practical adoption of AI in healthcare. Although there have been several efforts to explain time series data \citep{sood2022feature, bento2021timeshap, nayebi2023windowshap, gupta2022obesity}, few have attempted to explain EHR data in a predictive model setting.

To address these limitations, researchers have explored explainable models that generate both predictions and explanations within the same architectural framework through joint training. This allows the model to learn self-explainable representations without needing separate explanation models. Using expert-knowledge-driven clinical concepts (e.g., SOFA organ system scores in ICU mortality prediction) as units of explanation, learned via supervised auxiliary tasks, can provide more meaningful and plausible insights for clinicians. While attention mechanisms have been explored for intrinsic interpretability by highlighting influential input features, as seen in models like RETAIN \citep{choi2016retain} and DeepSOFA \citep{shickel2019deepsofa}, they often still focus on individual features rather than high-level concepts. The proposed HyMaTE (Mamba-based Architecture for Representation with Fusion-attention and Interpretation) model, designed for longitudinal data representation, explicitly aims to incorporate attention for interpretation, utilizing a fusion-attention mechanism to generate global attention weights and capture richer, more nuanced information, aligning with the broader goal of enhancing model transparency for clinical utility.

\begin{figure*}[htbp]
    \centering
    \includegraphics[width=0.95\linewidth]{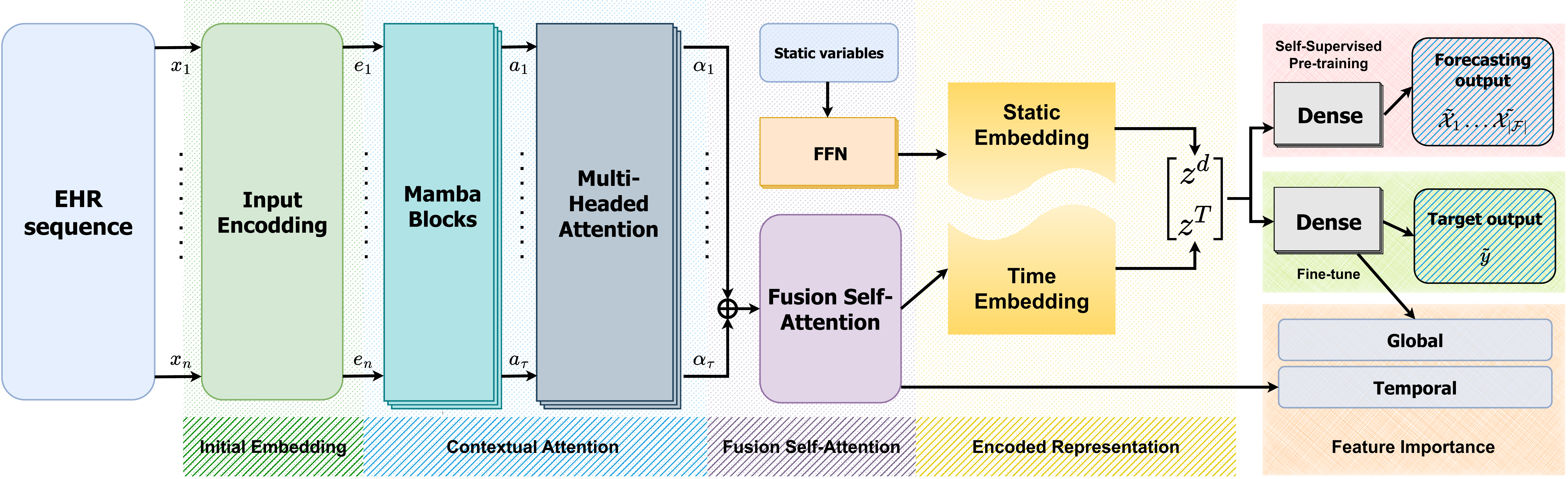}
    \caption{The proposed model architecture: HyMaTE}
    \Description{This figure presents the overall architecture of the proposed model for learning EHR representation. It contains several modules sequentially: Initial embedding, Contextual attention, Fusion self-attention, Encoded representation and Feature importance. The blocks are colored with bold black-faced types for names.}
    \label{fig:fig1}
\end{figure*}

\section{Methods}
We initially introduce the essential notations for problem definition, followed by an outline of the input data structure and the proposed model architecture. We then detail the self-supervised pre-training methodology, which utilizes masked forecasting, and the subsequent fine-tuning process for the downstream task. 

\subsection{Problem Definition}
Our model operates on a dataset structure typical of Electronic Health Records (EHR). Each patient's record comprises an irregular time series of observations, such as vital signs and laboratory results, alongside a collection of static variables like demographics that remain constant over time. This dataset, denoted as $D = {\{(\textbf{s}^k,\textbf{T}^k, y^k)\}}_{k=1}^N$, consists of $N$ labeled samples. For the $k$-th sample, $\textbf{s}^k \in \mathbb{R}^D$ represents a static vector of $D$ variables, $\textbf{T}^k=(t_1^k, t_2^k,\cdots,t_{n_k}^k)$ is a multivariate time-series of variable length $n_k$, and $y^k \in \{0,1\}$ is a binary label for a predefined task (e.g., mortality or condition). Each event $t_i^k$ is an observation triplet, formally defined as $(t, f, v)$, where $t \in \mathbb{R}_{\geq 0}$ is the observation time, $f \in \mathcal{F}$ is the feature or variable, and $v \in \mathbb{R}$ is the observed value (e.g., \texttt{[163.2 days, heart-rate, 68bpm]}). For a given patient (omitting the superscript $k$), a multivariate time-series $\textbf{T}$ of length $n$ is thus a set of $n$ such observation triplets: $\textbf{T} = \{{(t_i , f_i ,v_i)}\}_{i=1}^{n}$. The set of time-series variables, $\mathcal{F}$, encompasses various clinical measurements, including vital signs (e.g., temperature), lab measurements (e.g., hemoglobin levels, body weight), and input/output events (e.g., fluid intake, urine output). Consequently, for a patient $p$, the objective is to predict $y^p$ based on their static variables $\textbf{s}^p$ and multivariate time series $T^p$.

\subsection{Model Architecture} 
The architecture of our model HyMaTE is illustrated in Figure \ref{fig:fig1}. Initially, the EHR input is passed through an embedding layer where we encode the inputs in a sequence of triplets, detailed in the following subsection. The embedded inputs are passed through multiple Mamba blocks, followed by a number of self-attention layers. This constitutes a contextual embedding that is then transferred to the Fusion Self-attention layer for aggregation. The static variables are embedded using a Feed Forward Network (FFN). The final embedding of the EHR sequence is obtained by concatenating the temporal embeddings from the Fusion Self-attention layer and static variables. 

\subsubsection{Input Embedding} 
HyMaTE's design choices for EHR data processing begin by representing EHR events as triplets, each consisting of a time, feature, and value (e.g., $(t_i, f_i, v_i)$). An initial embedding maps each of these $n$ input triplets into a $\tau$-dimensional vector $e_i \in \mathbb{R}^\tau$. This is calculated by summing embeddings for the feature ($e_i^f$), continuous value ($e_i^v$), and time ($e_i^t$) components: $e_i = e_i^f + e_i^v + e_i^t \in \mathbb{R}^{d_t}$. The feature embedding uses a look-up table, while continuous value and time embeddings are derived using one-to-many Feed Forward Networks (FFNs) with learnable parameters: $e_i^v = \text{FFN}^v(v_i)$ and $e_i^t = \text{FFN}^t(t_i)$. This approach of representing EHR events as triplets has shown promise in earlier works \citep{horn2020set, tipirneni2022self}. 

\subsubsection{Contextual Attention} The input embeddings ${e_1, e_2, \dots, e_n}$ are then fed into the Contextual Attention module. This module comprises $M$ blocks, each containing $h$ Mamba layers. Each block takes $n$ input embeddings $\mathcal{E}\in \mathbb{R}^{n\times \tau}$ and applies normalization using root mean square normalization (RMS normalization). The tensor is then expanded through two linear projections. One projection undergoes a convolution followed by a Sigmoid Linear Unit (SiLU) activation, with its output processed by the discretized State-Space Model (SSM) to filter relevant information. The other projection is directly passed through a SiLU activation and then combined with the SSM outputs using a multiplicative gate. The combined output subsequently passes through another linear projection and is summed with the initial input $\mathcal{E}$, yielding the final output embeddings $\mathcal{C}_E\in \mathbb{R}^{n\times \tau}$. The mathematical formulations of the SSM unit are discussed below.
\paragraph{SSM unit} The Mamba block's core component is an SSM unit. This unit maps an input sequence $x(t)\in \mathbb{R}$ to an output sequence $y(t)\in \mathbb{R}$ via an implicit hidden state $h(t)\in \mathbb{R}^\psi$, where $\psi$ is the state size. The continuous-time linear differential equations defining this mapping are: \begin{equation} h'(t)=\textbf{A}h(t)+\textbf{B}x(t) \tag{1} \end{equation} \begin{equation} y(t)=\textbf{C}h(t) \tag{2} \end{equation} Here, $\textbf{A}\in {\mathbb{R}}^{\psi\times \psi}, \textbf{B}\in {\mathbb{R}}^{\psi \times 1},$ and $\textbf{C}\in {\mathbb{R}}^{1\times \psi}$ are learnable matrices. For multidimensional sequences, this system is applied independently to each dimension \citep{guefficiently}. To apply SSMs to discrete sequences, the system is discretized using a step size $\Delta$: \begin{equation} h_t=\overline{\textbf{A}}h_{t-1}+\overline{\textbf{B}}x_t \tag{3} \end{equation} \begin{equation} y_t=\textbf{C}h_t \tag{4} \end{equation} The discrete parameters $(\overline{\textbf{A}}, \overline{\textbf{B}})$ are derived using the zero-order hold (ZOH) rule \citep{gumamba, guefficiently} : \begin{equation} \overline{\textbf{A}}=\exp{(\Delta \textbf{A})} \tag{5} \end{equation} \begin{equation} \overline{\textbf{B}}={(\Delta \textbf{A})}^{-1}((\exp{(\Delta \textbf{A})-\textbf{I})\cdot \Delta \textbf{B} } \tag{6} \end{equation} The model employs a global convolution kernel $\overline{K}$ to enable parallel sequence processing and scaling during training, following the equations \citep{gumamba}: \begin{equation} \overline{K}=(C\overline{\textbf{B}}, C\overline{\textbf{AB}}, \dots, C\overline{\textbf{A}}^k\overline{\textbf{B}},\dots) \tag{7} \end{equation} \begin{equation} y=x *\overline{K} \tag{8} \end{equation}

\subsubsection{Self-Attention Layer} After the efficient sequence encoding performed by the Mamba components, self-attention layers are employed to extract local context from the sequences. These layers process the output from the Mamba encoding, capturing immediate dependencies and local patterns within the sequence. This local context extraction is a crucial intermediate step that precedes the global synthesis performed by the subsequent fusion-attention mechanism.

\subsubsection{Fusion Self-Attention} Following the contextual encoding and local context extraction, the output from the last Mamba block, $\mathcal{C}_E={c_1, c_2, \dots, c_n}$, is passed to the Fusion Self-attention module. Here, an FFN is applied to each contextual embedding $c_i$, and a softmax function is then used to calculate the attention weights, $\alpha_i$. The final embedding for the time-series, $z^T$, is computed by a weighted sum of these contextual triplet embeddings and their corresponding attention weights: \begin{equation}
    a_i=\textbf{u}_a^T \tanh{(\textbf{W}_a \textbf{c}_i +\textbf{b}_a)}
\end{equation}
\begin{equation}
    \alpha_i={\frac{\exp{(a_i)}}{\sum_{j=1}^\tau \exp{(a_j)}} \hspace{2em}} \forall i=1,\dots,\tau
\end{equation}
\begin{equation}
    z^T=\sum^\tau_{i=1} \alpha_i \textbf{c}_i
\end{equation}
where $\textbf{W}_a\in \mathbb{R}^{d_a\times \tau}, \textbf{b}_a\in \mathbb{R}^{d_a},\textbf{u}_a\in \mathbb{R}^{d_a}$ are the weights of this attention network which has $d_a$ neurons in the hidden layer.
For the demographic variables, the embedding, $z^d$, is generated by passing $s\in \mathbb{R}^D$, a vector of $D$ static variables, forwarding through an FFN with 2 hidden layers and an output layer of $d$ nodes, where $d$ is the hyperparameter controlling the size of the embedded demographic variable vector. 
\begin{equation}
    z^d=\tanh(\textbf{W}_2^d \tanh(\textbf{W}_1^d \textbf{d}+\textbf{b}_1^d)+\textbf{b}_2^d)\in\mathbb{R}^d
\end{equation}where the hidden layer has a dimension of $2d$.
Thus, we encode variable-length time-series to a fixed-length vector, $e^E$, concatenating demographic variable embeddings $z^d$ and time variable embeddings $z^T$,
\begin{equation}
e^E = \begin{bmatrix}
z^d \\ z^T
\end{bmatrix} \in \mathbb{R}^{d_E}
\end{equation} where $d_E = \tau + d$ represents the sum of dimensions for time embeddings and demographic embeddings.

\subsection{Training} HyMaTE is trained using a semi-supervised approach, which involves two phases: self-supervised pre-training, followed by supervised fine-tuning on downstream tasks.

\subsubsection{Self-supervised pre-training} The model incorporates forecasting as a self-supervision task during pre-training to learn robust representations from limited labeled EHR data. This phase uses a larger dataset $D'={{(\textbf{s}^k, \textbf{T}^k, \textbf{m}^k, \mathcal{X}^k)}}_{k=1}^{N'}$, where $N' \ge N$ samples are available. Here, $\textbf{m}^k \in {\{0,1\}}^{|\mathcal{F}|}$ is the forecast mask indicating whether each variable was observed in the forecast window, and $\mathcal{X}^k \in \mathbb{R}^{|\mathcal{F}|}$ contains the corresponding variable values when observed, with $\mathcal{F}$ being the underlying set of time-series variables. 

The forecast mask is used for excluding unobserved forecasts during training, ensuring that only relevant data informs the loss function. The forecasting task utilizes the learned patient representation $\tilde{\mathcal{X}}$, obtained by passing the concatenated embeddings through a dense layer with weights $w_s \in \mathbb{R}^{|\mathcal{F}|}$ and bias $b_s \in \mathbb{R}$:\begin{equation} \tilde{\mathcal{X}} = w_s\begin{bmatrix} z^d \ \\ z^T \end{bmatrix}+b_s \in \mathbb{R}^{|\mathcal{F}|} \tag{13} \end{equation}Masked Mean Squared Error (MSE) loss is used for training on the forecasting task to account for missing values in the forecast outputs. The self-supervised forecasting loss is given by:\begin{equation} \mathcal{L}_{ss}=\frac{1}{|N'|}\sum^{N'}_{k=1}\sum^{|\mathcal{F}|}_{j=1}{m_j^k{(\tilde{\mathcal{X}}_j^k-\mathcal{X}^k_j)}^2} \tag{14} \end{equation} where $m_j^k \in {0,1}$ indicates the availability (1) or unavailability (0) of the ground truth forecast $\mathcal{X}^j_k$ for the $j$-th variable in the $k$-th sample.

\subsubsection{Fine-tuning} During the fine-tuning phase, the patient representation $e^E$ extracted from the fusion self-attention layer is used, and heads tailored to the downstream tasks are attached. The final prediction for the target task is obtained by passing the concatenated embeddings through a dense layer with weights $w_o^T\in \mathbb{R}^{d_E}, b_o\in \mathbb{R}$ and a sigmoid activation function.
\begin{equation}
    \tilde{y}=\sigma(w_o^T\begin{bmatrix} z^d \ \\ z^T \end{bmatrix}+b_o)
\end{equation}

\section{Experiments}
To rigorously demonstrate the efficacy and generalizability of HyMaTE, our model was extensively evaluated on diverse EHR datasets, encompassing both public and real-world private clinical data. The experimental setup involved specific data preprocessing, task definitions tailored to each dataset, and an innovative approach to evaluating model interpretability.

    \subsection{Datasets}
    HyMaTE's performance was evaluated using three EHR datasets, showcasing its robustness and adaptability in diverse environments. 

        \subsubsection{PhysioNet Challenge 2012 \citep{silva2012predicting}}
        The PhysioNet Challenge 2012 dataset is a publicly available, standardized resource designed for predicting in-hospital mortality within ICU stays, with $14.2\%$ of positive labels. The dataset consists of $11,988$ patients with $42$ different variables, including $37$ time series event-types. Data preprocessing aligns with the methodology described in \citep{tipirneni2022self} for mortality prediction. Observation windows are defined as the first 48 hours of ICU stay and for predicting mortality within the subsequent $2$-hour period. Data from sets B and C are combined and split into training and validation sets ($80\%:20\%$), with set A reserved for testing.
        
        \subsubsection{MIMIC-IV \citep{johnson2020mimic}}
        The MIMIC-IV dataset is a publicly available, de-identified collection of retrospective patient data from the Beth Israel Deaconess Medical Center. Our evaluation focuses on a derived ICU subset, comprising $53,150$ patients and $69,211$ admissions. Following the methodology of \citep{labach2023duett}, we define the mortality prediction task within this ICU dataset. The initial $48$ hours of a patient's stay serve as the input time window, with the objective of predicting subsequent in-hospital mortality, length of stay exceeding 1 week and readmission within 1 month. A patient-level split of $60\%:20\%:20\%$ is employed for the training, validation, and test sets, respectively.
        
        \subsubsection{Pediatric weight management}
        The Pediatric Weight Management dataset was extracted from the Electronic Health Records (EHR) of a large pediatric healthcare system, encompassing primary, specialty, inpatient, and emergency care across five U.S. states. The de-identified EHR data is structured using a pediatric-specific Common Data Model (CDM), which anchors patient encounters within the health system. This CDM is based on the Observational Medical Outcomes Partnership (OMOP) model \citep{makadia2014transforming}. The target task involves predicting a $5\%$ maximum weight loss in patients utilizing anti-obesity medications. This dataset comprises $14,392$ individual pediatric records, with $18.7\%$ positive labels and a maximum time series length of $8,654$ observations.

    \subsection{Baseline Models}
    From a total of nine baseline models evaluated, the top four performers, based on their superior performance metrics, are presented in the following discussion. A comprehensive table detailing the results of all ten baseline models, including their full performance statistics, is provided in Appendix A. 
        \subsubsection{Set Functions for Time Series (SeFT)} \citep{horn2020set} This model is designed to process a set of observation triplets. It incorporates sinusoidal encoding to effectively embed temporal information. Additionally, the deep network employed to integrate the observation embeddings is structured as a set function, utilizing a simplified yet efficient variation of multi-head attention.
        \subsubsection{Self-supervised Transformer for Time-Series (STraTS)} \citep{tipirneni2022self}: The model uses a continuous value embedding technique to encode continuous time and variable values without discretization. It features a Transformer component with multi-head attention layers, allowing it to learn contextual triplet embeddings while avoiding issues related to recurrence and vanishing gradients found in recurrent architectures.
        \subsubsection{EHR-Mamba} \citep{fallahpour2024ehrmamba}  This model encodes EHR data in a combination of seven different embeddings and uses stacked Mamba blocks for mapping input sequence to output tensor for downstream forecasting or predicting tasks.
        \subsubsection{DuETT} \citep{labach2023duett} This architecture extends transformers to exploit both time and event modalities of EHR data. Each DuETT layer is made up of two Transformer sub-layers that attend along the event and time dimensions, respectively.  The representation learned from the time and event embeddings is used for downstream tasks by attaching respective task heads.

    \subsection{Implementation Details} 
    Model training was conducted with a fixed batch size of $32$ utilizing the Adam optimizer. Training was terminated if the validation performance on evaluation metrics did not improve for $10$ consecutive epochs, based on a predefined patience threshold. For all models, the primary evaluation metrics employed were $(i)$ AUROC: Area Under the Receiver Operating Characteristic curve, and $(ii)$ AUPRC: Area Under the Precision-Recall curve. All model implementations were adapted to PyTorch modules from their respective official codebases. Data preprocessing and general customizations were derived from the publicly available codebase \url{http://github.com/sindhura97/STraTS}. All experiments were performed on a single Tesla T4 tensor core GPU hosted on the Amazon Web Services (AWS) SageMaker platform. Our implementations are publicly available at \url{https://github.com/healthylaife/HyMaTE}.

\begin{table*}[htbp]
\renewcommand{\arraystretch}{1.2}
\centering
\caption{Performance Comparison of HyMaTE against baseline models on 5 clinical tasks using 3 EHR datasets. The results show the mean and standard deviation of the metrics with 10 repeated experiments. The best results are highlighted in boldface, and the second-best results are underlined.}
\begin{tabular}{c *{5}{c}}
\toprule
Metric & SeFT \citep{horn2020set} & STraTS \citep{tipirneni2022self} & EHR-Mamba \citep{fallahpour2024ehrmamba} & DuETT \citep{labach2023duett} & HyMaTE (Ours) \\
\midrule
\multicolumn{6}{c}{PhysioNet 2012 : mortality prediction} \\
\hline
AUROC & 0.812 \(\pm\) 0.008 & 0.838 \(\pm\) 0.009 & 0.844 \(\pm\) 0.006 & \underline{0.857 \(\pm\) 0.018} & \textbf{0.868 \(\pm\) 0.045} \\
AUPRC & 0.464 \(\pm\) 0.012 & 0.487 \(\pm\) 0.001 & 0.534 \(\pm\) 0.026 & \underline{0.598 \(\pm\) 0.056} & \textbf{0.602 \(\pm\) 0.015} \\
\hline
\multicolumn{6}{c}{MIMIC-IV : mortality prediction} \\
\hline
AUROC & 0.644 \(\pm\) 0.007 & 0.692 \(\pm\) 0.002 & \underline{0.881 \(\pm\) 0.021} & 0.852 \(\pm\) 0.007 & \textbf{0.907 \(\pm\) 0.025} \\
AUPRC & 0.288 \(\pm\) 0.001 & 0.295 \(\pm\) 0.002 & \underline{0.637 \(\pm\) 0.095} & 0.605 \(\pm\) 0.048 & \textbf{0.648 \(\pm\) 0.021} \\
\hline
\multicolumn{6}{c}{MIMIC-IV : length of stay prediction} \\
\hline
AUROC & 0.687 \(\pm\) 0.064 & 0.732 \(\pm\) 0.095 & \underline{0.781 \(\pm\) 0.027} & 0.778 \(\pm\) 0.048 & \textbf{0.802 \(\pm\) 0.055} \\
AUPRC & 0.312 \(\pm\) 0.023 & 0.358 \(\pm\) 0.048 & 0.428 \(\pm\) 0.015 & \underline{0.435 \(\pm\) 0.038} & \textbf{0.495 \(\pm\) 0.011} \\
\hline
\multicolumn{6}{c}{MIMIC-IV : readmission prediction} \\
\hline
AUROC & 0.638 \(\pm\) 0.042 & 0.718 \(\pm\) 0.028 & 0.752 \(\pm\) 0.024 & \underline{0.772 \(\pm\) 0.024} & \textbf{0.798 \(\pm\) 0.035} \\
AUPRC & 0.377 \(\pm\) 0.004 & 0.544 \(\pm\) 0.022 & 0.585 \(\pm\) 0.024 & \underline{0.592 \(\pm\) 0.024} & \textbf{0.624 \(\pm\) 0.032} \\
\hline
\multicolumn{6}{c}{Pediatric : weight-loss prediction} \\
\hline
AUROC & 0.644 \(\pm\) 0.007 & 0.692 \(\pm\) 0.002 & \underline{0.704 \(\pm\) 0.006} & 0.654 \(\pm\) 0.008 & \textbf{0.752 \(\pm\) 0.014} \\
AUPRC & 0.288 \(\pm\) 0.001 & \underline{0.295 \(\pm\) 0.002} & 0.284 \(\pm\) 0.012 & 0.257 \(\pm\) 0.051 & \textbf{0.387 \(\pm\) 0.018} \\
\bottomrule
\end{tabular}
\label{tab:tab1}
\end{table*}

\section{Results}
HyMaTE demonstrated consistent performance across various clinical predictive tasks on public EHR datasets, including PhysioNet Challenge 2012 and MIMIC-IV, as well as the private pediatric clinical dataset, presented in Table \ref{tab:tab1}. Comprehensive ablation experiments demonstrated that HyMaTE outperforms existing Transformer-based and standalone Mamba models, as shown in Table \ref{tab:tab2}.

For mortality prediction on the PhysioNet 2012 dataset, HyMaTE achieved an AUROC of $0.868$ and an AUPRC of $0.602$. HyMaTE's performance was notably strong, closely following DuETT, securing the second-best results in both AUROC of $0.857$ and AUPRC of $0.598$. On the MIMIC-IV dataset, HyMaTE secured the best performance for mortality prediction with an AUROC of $0.907$ and an AUPRC of $0.648$, representing an improvement of $0.026$ in AUROC and $0.011$ in AUPRC over the second-best EHR-Mamba (AUROC $0.881$, AUPRC $0.637$). For length of stay prediction on MIMIC-IV, HyMaTE again achieved best performance, securing the best AUROC of $0.802$ and the best AUPRC of $0.495$, exceeding EHR-Mamba's AUROC of $0.781$ by $0.021$ and its AUPRC of $0.428$ by $0.067$. In readmission prediction on MIMIC-IV, HyMaTE maintained its lead with the best AUROC of $0.798$ and AUPRC of $0.624$, outperforming DuETT's AUROC of $0.772$ by $0.026$ and its AUPRC of $0.592$ by $0.032$. Furthermore, on the private pediatric clinical dataset for weight-loss prediction, HyMaTE achieved the best AUROC of $0.752$ and AUPRC of $0.387$, significantly surpassing EHR-Mamba, which recorded an AUROC of $0.704$ and an AUPRC of $0.284$. The consistently strong results demonstrate HyMaTE's efficiency and capability to process longer sequences efficiently, positioning it as a scalable and generalizable solution for real-world healthcare applications.

\begin{table*}[htbp]
    \centering
    \caption{Predictive performance analysis with ablations (AUROC).}
    \begin{tabular}{l|c|c|c|c|c}
        \toprule
        \multirow{2}{*}{Ablations} & \multirow{2}{2.0cm}{\centering PhysioNet 2012 mortality}  & \multicolumn{3}{c}{MIMIC-IV}& \multirow{2}{2.0cm}{\centering Pediatric weight loss}\\
 & & mortality& length of stay& readmission& \\
        \midrule
        w/o Input embedding & $0.815 \pm 0.024$ & $0.847 \pm 0.035$  & $0.734 \pm 0.015$ & $0.724 \pm 0.011$ & $0.718 \pm 0.027$ \\ \hline
        w/o Self-Attention layer & $0.855 \pm 0.021$ & $0.861 \pm 0.014$  & $0.764 \pm 0.028$ & $0.738 \pm 0.045$ & $0.724 \pm 0.026$ \\ \hline
        w/o Attention fusion & $0.842 \pm 0.016$ & $0.868 \pm 0.021$  & $0.758 \pm 0.012$ & $0.747 \pm 0.035$ & $0.731 \pm 0.017$ \\ \hline
        w/o Self-supervised Pretraining & $0.784 \pm 0.035$ & $0.812 \pm 0.038$  & $0.741 \pm 0.012$ & $0.722 \pm 0.021$ & $0.705 \pm 0.015$ \\ \hline
        w/o Mamba blocks & $0.806 \pm 0.024$ & $0.817 \pm 0.04$  & $0.725 \pm 0.048$ & $0.715 \pm 0.015$ & $0.718 \pm 0.027$ \\ \hline
        \textbf{HyMaTE (ours)} & $\textbf{0.868} \pm \textbf{0.045}$  & $\textbf{0.907} \pm \textbf{0.025}$  & $\textbf{0.802} \pm \textbf{0.055}$  & $\textbf{0.798} \pm \textbf{0.035}$ & $\textbf{0.752} \pm \textbf{0.014}$ \\
        \bottomrule
    \end{tabular}
    \label{tab:tab2}
\end{table*}

\subsection{Ablation Study}
To demonstrate the contribution of each architectural component to HyMaTE's performance, we conducted an ablation study, systematically removing key elements and evaluating the resultant performance drop in terms of AUROC across the three clinical datasets. The full HyMaTE model achieved baseline AUROC scores. The results of this study are comprehensively detailed in Table \ref{tab:tab2}.

Removing the input embedding layer ("w/o Input embedding") led to a notable decrease in performance across all datasets. This significant degradation highlights the critical importance of HyMaTE's initial embedding module. The observed performance drop confirms that this initial, context-aware representation is fundamental for the model to effectively capture the complexities of raw EHR data, inherently diminishing the effect of missingness in time series data by allowing for adaptive representations. 

Our ablation study showed that removing either the self-attention layer or the attention fusion mechanism resulted in identical performance drops across all datasets. This suggests an interdependent relationship between these two components within HyMaTE's hybrid attention architecture. This consistent decline underscores the necessity of HyMaTE's advanced attention mechanisms for enhancing feature blending and capturing richer representations. While Mamba excels at sequence mixing, the self-attention layers are designed to extract local context, and the fusion-attention mechanism generates global attention weights, synthesizing insights into a unified representation. This multi-layered attention approach is vital for HyMaTE to process complex EHR data effectively.

Removing the self-supervised pretraining phase had a significant negative impact on the model's performance across all evaluated datasets, especially on the Pediatric Weight Dataset. This notable decline highlights the importance of the semi-supervised training approach employed by HyMaTE. The self-supervised forecasting task during pretraining allows the model to learn strong representations from the limited labeled EHR data available. This is particularly advantageous in healthcare, where standardized labeled EHR datasets are rare and privacy concerns are critical. The results indicate that this pretraining step is essential for HyMaTE to develop generalizable, high-quality representations, making the model resilient to data limitations and vital for tackling complex tasks.

\begin{figure}[htbp]
    \centering
    \includegraphics[width=0.95\linewidth]{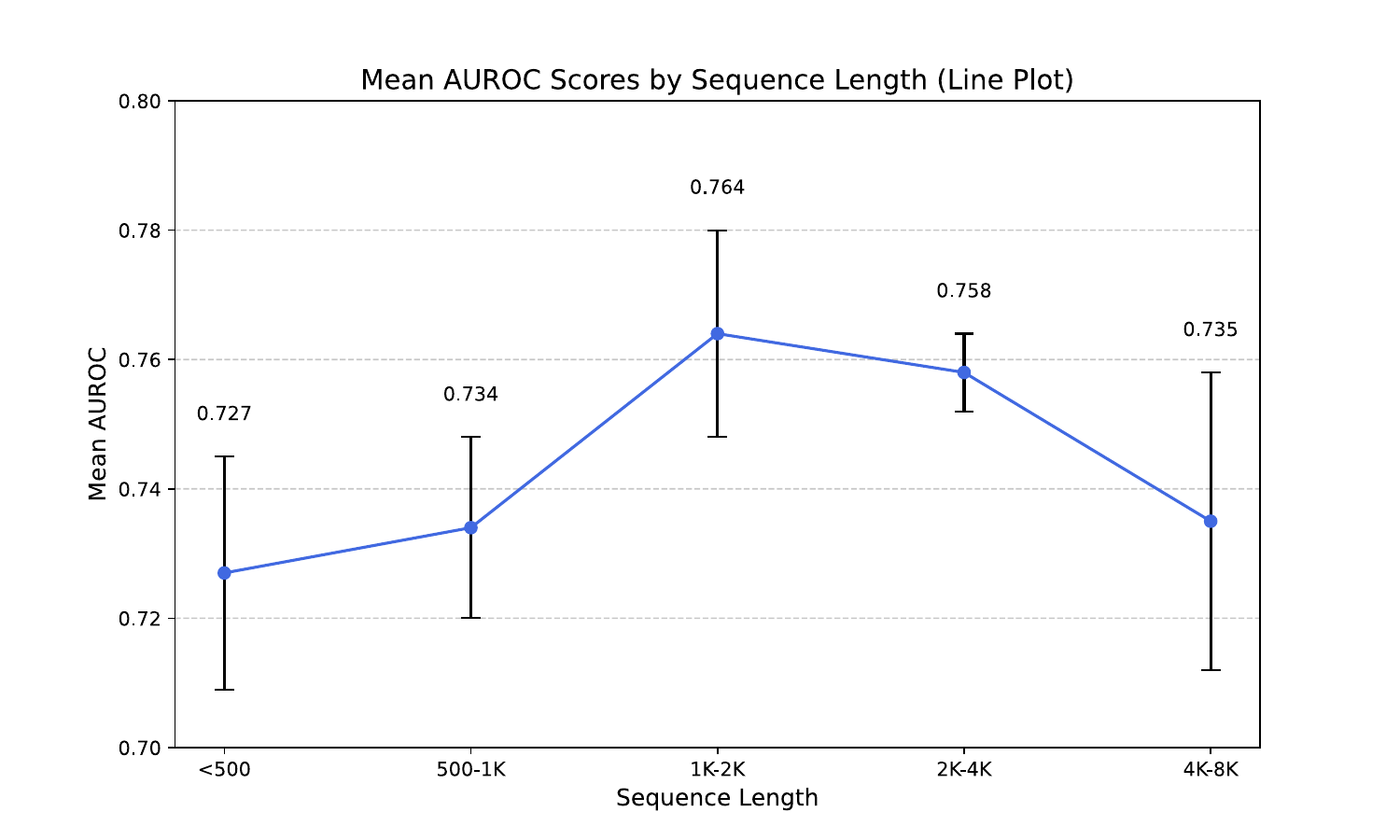}
    \caption{Mean AUROC vs sequence length}
    \label{fig:fig2}
\end{figure}

Besides removing the Mamba blocks for the ablation study, drastically reducing the performance, the integration of the Mamba module has positively impacted inference times by streamlining the contextual embedding process. It’s worth mentioning, however, that the hybrid model's performance may not surpass that of the standalone Mamba block model. Additionally, the inclusion of Mamba supports the capability to utilize longer contexts effectively. We present the mean AUROC for weight loss prediction in the pediatric dataset, classified by sequence lengths in Figure \ref{fig:fig2}. It is important to highlight that, despite a limited number of samples in the 4K to 8K length range, the decline in AUROC performance was minimal.

\begin{table}
    \centering
    \caption{Global feature importance: PhysioNet 2012 mortality prediction task. [NIMAP: Non-invasive mean arterial blood pressure, HR: Heart rate, BUN: Blood urea nitrogen, NISysABP: Non-invasive systolic arterial blood pressure]}
    \begin{tabular}{l|c}
    \toprule
         Variable& Contribution score\\
         \midrule
         NIMAP & 4.481\\
         HR     & 3.222\\
         BUN & 1.355\\
         NISysABP& 0.492\\
         Temp & 0.217\\
         ICUType\textunderscore3  & 0.139\\
         Age  & 0.029\\
         Urine& 0.018\\
         \bottomrule
    \end{tabular}
    
    \label{tab:tab3}
\end{table}

\subsection{Interpretability}

Interpretability of our model's predictions is achieved through the analysis of attention weights derived from the fusion-attention layer and the coefficients of the final predictive dense layers, providing both temporal and global insights.

For temporal interpretability, attention weights corresponding to temporal variables were extracted across all timestamps for each individual patient. This process allowed for the calculation of individual attention scores for each variable at every timestamp. This granular analysis facilitates an understanding of the model's decision-making process for a given target label, specifically identifying the critical segments within the time series that influenced the prediction. To illustrate this temporal interpretability, a case study is presented involving an 82-year-old male patient from the PhysioNet-2012 dataset, who experienced in-hospital mortality on the fourth day of ICU admission. For this patient, our model predicted a probability of in-hospital mortality of 0.935. A temporal interpretation of the model's attention weights revealed a heightened focus on timestamps corresponding to the collection of Non-invasive Mean Arterial Blood Pressure (NIMAP, mmHg) data. Fluctuations in NIMAP, among other temporal variables, have previously been associated with increased ICU mortality \citep{shao2022u}.

Global feature importance was determined by analyzing the weights of the final dense layer, which yielded a ranked list of variables contributing to the overall predictive model. Table \ref{tab:tab3} presents the top eight contributing variables identified in the mortality prediction task using the PhysioNet 2012 dataset. These variables align with established clinical findings, having been consistently associated with ICU mortality and patient deterioration in prior research \citep{giri2022blood, lyons2019factors, deng2025association, sarkar2022mean, palacios2023mean, shao2022u}.

\section{Discussion}

The core contribution of this work is HyMaTE, a novel architecture designed to overcome the inherent complexities of EHR. While standalone Transformer models are constrained by quadratic computational complexity and limited context lengths, and traditional SSMs like Mamba primarily focus on sequence mixing, HyMaTE strategically amalgamates the strengths of both Mamba and Transformer's attention mechanisms to provide a more robust and efficient solution.

HyMaTE's hybrid design directly addresses the limitations of its predecessors by leveraging complementary strengths. It integrates Mamba blocks for efficient linear-time sequence encoding, enabling the processing of extensive EHR data sequences with substantially reduced computational costs. This is crucial for scalability in real-world healthcare applications, where patient histories can span thousands of tokens. To address Mamba's limitation in channel mixing—a critical element for multivariate time series data like EHRs, where inter-feature relationships are essential—HyMaTE incorporates advanced attention mechanisms found in Transformer models. Following the initial Mamba encoding, self-attention layers are employed to extract local context, capturing immediate dependencies and intricate patterns within the sequence. Subsequently, a fusion self-attention mechanism generates global attention weights, synthesizing insights from both local and global contexts into a unified representation. This synergistic combination of Mamba's efficient sequence processing with the Transformer's powerful attention for channel mixing and contextual understanding allows HyMaTE to capture both long-range dependencies and fine-grained inter-variable relationships within the EHR.

A critical aspect contributing to HyMaTE's robustness is its semi-supervised training methodology, which incorporates a self-supervised forecasting task during the pre-training phase. This pre-training is essential as it enables the model to learn robust representations from limited labeled EHR data, offering a significant advantage given the scarcity of standardized labeled EHR datasets and pervasive privacy concerns in healthcare. Ablation studies conclusively demonstrated a considerable negative impact on performance when this pre-training phase was omitted, underscoring its fundamental role in enabling HyMaTE to learn generalizable and high-quality representations, making it robust against data limitations and crucial for complex predictive tasks.

HyMaTE's consistent superior performance across various clinical prediction tasks, evaluated on diverse EHR datasets including PhysioNet Challenge 2012, MIMIC-IV, and a proprietary pediatric clinical dataset, demonstrates the effectiveness of its hybrid design. Beyond its predictive capabilities, HyMaTE explicitly aims for interpretability, achieved through the analysis of attention weights derived from the fusion-attention layer and the coefficients of the final predictive dense layers. This provides both temporal and global insights, facilitating the identification of critical temporal segments and globally important features that influence predictions, thereby aligning with established clinical findings and enhancing model transparency for practical clinical utility.

These results suggest that HyMaTE offers a promising avenue for advancing EHR representation learning, leading to more accurate, efficient, and interpretable predictive models that can support clinical decision-making and ultimately improve patient outcomes. Future directions for this research include exploring the integration of other advanced SSM variants or novel attention mechanisms to further optimize performance and efficiency. Additionally, evaluating HyMaTE's capabilities in real-time clinical prediction scenarios and investigating its applicability to a broader range of clinical tasks, such as patient phenotyping or treatment recommendation, would be valuable. Further efforts will also focus on enhancing the interpretability tools to facilitate more direct and actionable insights for clinicians, and on rigorously assessing the model's fairness and generalizability across diverse patient populations and healthcare systems.

\section{Conclusion}
In conclusion, HyMaTE presents a significant advancement in EHR representation learning through its innovative hybrid architecture, effectively combining the linear-time efficiency of Mamba with the powerful contextual and channel-mixing capabilities of Transformer attention mechanisms. Its robust performance across diverse clinical prediction tasks and datasets, coupled with its inherent interpretability and semi-supervised training approach, underscores its potential to generate accurate, efficient, and transparent predictive models. HyMaTE's ability to derive meaningful insights from complex and often incomplete EHR data holds promise for enhancing clinical decision support and ultimately contributing to improved patient care in real-world healthcare settings.

\begin{acks}
Our study was supported by NSF award 2443639, NIH awards, P20GM103446, and
U54-GM104941. We also acknowledge computational credits from Amazon Web Services (AWS).
\end{acks}

\bibliographystyle{ACM-Reference-Format}
\bibliography{bibliography}

\end{document}